\documentclass{article}

 \PassOptionsToPackage{numbers, compress}{natbib}
\usepackage[final]{nips_2018}

\usepackage{amsmath,amsfonts,bm}

\def\eqref#1{equation~\ref{#1}}
\def\1{\bm{1}}

\DeclareMathAlphabet{\mathsfit}{\encodingdefault}{\sfdefault}{m}{sl}
\SetMathAlphabet{\mathsfit}{bold}{\encodingdefault}{\sfdefault}{bx}{n}

\usepackage[utf8]{inputenc} % allow utf-8 input
\usepackage[T1]{fontenc}    % use 8-bit T1 fonts
\usepackage{hyperref}       % hyperlinks
\usepackage{url}            % simple URL typesetting
\usepackage{booktabs}       % professional-quality tables
\usepackage{amsfonts}       % blackboard math symbols
\usepackage{nicefrac}       % compact symbols for 1/2, etc.
\usepackage{microtype}      % microtypography
\usepackage{algorithm2e}
\usepackage{algcompatible}
\usepackage{amssymb}
\usepackage{graphicx}
\usepackage{titlesec}
\usepackage{amsmath}
\usepackage{arydshln}
\usepackage[position=bottom]{subfig}
\usepackage{wrapfig}
\usepackage{enumitem}
\usepackage{multirow}
\usepackage{adjustbox}

\DeclareGraphicsExtensions{%
    .png,.PNG,%
    .pdf,.PDF,%
    .jpg,.mps,.jpeg,.jbig2,.jb2,.JPG,.JPEG,.JBIG2,.JB2}
\def\L{{\cal L}}

\titlespacing*{\section}{0pt}{-0.5pt}{-0.5pt}
\titlespacing*{\subsection}{0pt}{-0.5pt}{-0.5pt}
\title{Towards Continuous Domain adaptation for Healthcare}
\author{ Rahul Venkataramani, Hariharan Ravishankar,
Saihareesh Anamandra\\
GE Global Research, Bangalore \\
\texttt{\{firstname.lastname@ge.com\}}
}

\def\memoryModule{\mathcal{M}}

\begin{document}

\maketitle

\begin{abstract}
Deep learning algorithms have demonstrated tremendous
success on challenging medical imaging problems.
However, post-deployment, these algorithms are susceptible
to data distribution variations owing to  \emph{limited data issues} and
\emph{diversity} in medical images. In this paper, we propose
\emph{ContextNets}, a generic memory-augmented neural network
framework for semantic segmentation to achieve continuous
domain adaptation without the necessity of retraining.
Unlike existing methods which require access to entire
source and target domain images, our algorithm can adapt to a target domain with a  few similar images.
We condition the inference on any new
input with features computed on its support set of images
(and masks, if available) through contextual embeddings
to achieve site-specific adaptation.
We demonstrate state-of-the-art domain adaptation performance on
 the X-ray lung segmentation problem from
three independent cohorts that differ in disease type, gender,
contrast and intensity variations.
\end{abstract}

% All individual source files BEGIN

\section{Introduction}
For successful translation of impressive lab results to clinically accepted products, deep learning algorithms have to overcome problems of domain adaptation \cite{kamnitsas2017unsupervisedShort} \cite{ghafoorian2017transferShort} and customization.
Data domain shift is routinely encountered in clinical settings due to variations in equipment, protocol, demography and pathological conditions.
Additionally, algorithms are expected to be customizable for a specific site/doctor to account for local preferences/guidelines.

A potential design solution to overcome above problems is to enable lifelong
learning from incremental corrective annotations, post-deployment.
The naive approach  of fine-tuning the model  on new domain data is infeasible without a large number of annotated samples. Further,
this leads to \emph{catastrophic forgetting}\cite{kirkpatrick2017overcoming} - phenomenon of  forgetting past knowledge when retrained on new data. To circumvent obtaining costly annotations, unsupervised domain adaptation approaches
transform the image from the target domain to match images from the source
domain. %, thereby resisting performance degradation.
Early attempts included image preprocessing steps like contrast
normalization, filtering etc. In \cite{kamnitsas2017unsupervisedShort}, authors employed an adverserial loss to ensure alignment of
feature embeddings from source  \& target domains. Recently, \cite{zhang2018task} used a variant of generative adverserial network (GAN), a CycleGAN to style transfer an image
from target  to source domain.
However, there are practical challenges:
1) Training a GAN at each deployment site for target to source mapping.
2) Access to source domain images/masks during deployment.
3) Availability of a large target domain cohort.
The assumption of mapping any new data to the source manifold may not hold true while encountering rare events/unseen target distributions.

Memory augmented neural networks (MANN) have recently emerged as a popular framework for classification problems. By augmenting fuzzy matching abilities of neural networks
with an external memory, MANNs have demonstrated ability to learn from few samples, remember  rare events \cite{2017arXiv170303129K} etc. Briefly,  for every new sample, features are retrieved by querying
the stored memory and are combined with the primary neural network  for
final prediction. % \ref{fig:mann}.
However, we are not aware of any attempt to use MANNs for continuous
domain adaptation or for semantic segmentation.

% MANNS, NO RETRAINING, FEW SAMPLES, LIFELONG, NO SOURCE DATA, VARIANTS - supervised/Unsupervised, EXPLAIN FIGURE, Contextual EMBEDDINGS

% \begin{wraptable}{l}{9cm}
%     \centering
%     \includegraphics[width=.65\textwidth]{Images/memoryDiagramWithEncoder.jpg}
%     \caption{Schematic for \emph{ContextNet}}
%     \label{fig:mann}
% \end{wraptable}

We propose \emph{ContextNets} - a lifelong learning methodology for continuous domain adaptation of any deep learnt semantic segmentation algorithm. %to any new target domain.
Compared to other methods, our approach:
1) Does not require any training post-deployment.
2) Does not require access to source domain data.
3) Is a single framework to perform domain adaptation in unsupervised/supervised variants.

\section{Methodology}
In this section we present the generic formulation and framework for continual domain adapation, while in Section \ref{Exp}, we present the practical realization of the same.

\subsection{Continual Adaptation via Memory}

The central part of the proposed approach is a memory module $\memoryModule^d$ , for every target domain $d,$ which aids in dynamically adapting to the domain distribution variations while observing new cases continually. Each row in $\memoryModule^d$ (Eq. \ref{Md}) corresponds to one sample image observed so far and contains two sets of information - \emph{Image Features} and \emph{Context Features}.

\textbf{Image Features}: Intuitively, the image features computed on image $I_k$, are used as keys to retrieve contextually similar images from domain $d$, while context features $c_1, c_2, .. c_n$ encapsulate the description of various properties of these images. $q$ and $c_i$ can be computed from respective query and context mapping functions as shown in Eq. \ref{FunctionMappers}.
%\begin{equation}
%\memoryModule^d = \{q, c_1, c_2, .. c_n\}
%\label{Md}
%\end{equation}
%\begin{equation}
%\boldsymbol{Q} :\boldsymbol{I_k}\rightarrow[q],
%\boldsymbol{C_i} :\boldsymbol{I_k}\rightarrow[c_i]
%\label{FunctionMappers}
%\end{equation}
\begin{gather}
\memoryModule^d = \{q, c_1, c_2, .. c_n\}
\label{Md}
\\
{Q} :{I_k}\rightarrow[q],
{C_i} :{I_k}\rightarrow[c_i]
\label{FunctionMappers}
\end{gather}

\textbf{Context Features}: For every test image $I_{k}$, this memory is used to return a support set $C(I_{k})$ by computing $T$ nearest neighbors on the query features (Eq. \ref{NN1}).
%The nearest neighbor operator $NN$ is defined by: $A = NN(I_{k}$, $\mathcal{M}) = argmax_{i} q(I_{k}) \cdot$  $(\mathcal{M}[q]_{i}) $.
Further, we also collect the \emph{context features} corresponding to these support set images (Eq. \ref{NN2}). We refer to support set of images $C(I_k)$ as \emph{Context Set} and context features of the support set  $C_v(I_k)$ as \emph{Context Vector}.
%\begin{equation}
%C(I_k) = \{NN( q(I_{k}), \mathcal{M}[q])\}_T
%\label{NN1}
%\end{equation}
%
%\begin{equation}
%C_v(I_k) = \mathcal{M}[c_1, c_2, .. c_n][C(I_k)]
%\label{NN2}
%\end{equation}
%
\begin{gather}
C(I_k) = \{NN( q(I_{k}), \mathcal{M}[q])\}_T
\label{NN1}
\\
C_v(I_k) = \mathcal{M}[c_1, c_2, .. c_n][C(I_k)]
\label{NN2}
\end{gather}

% Query functions can be designed based on image similarity, or subject information like gender, demography, age, disease diagnosis, etc. Similarly, context mapping functions can designed to model common properties like texture, contrast, shape properties of masks, etc. In section \ref{XrayDA}, we describe the design choices made for lung segmentation in X-ray.

\subsection{Contextual Embedding}
\label{formulation}
A fully convolutional network  ((FCN \cite{ronneberger2015u}) is a mapping $D\circ\ E$, where the {\em encoder} $E$ and {\em decoder} $D$ are parameterized by a set of parameters $\theta$. Given training pairs of images and segmentation masks $\{I_k,S_k\}$, $k=1,2,\hdots,N$, one learns the parameters $\theta$ by minimizing the training loss $J(\theta) = \sum_{k=1}^N \mathcal{L}(S_{k}, \hat{S}_{k}(\theta) ) $, e.g. RMSE loss, binary cross-entropy loss (BCE), where  $\hat{S}_{k}(\theta)=(D\circ E)[I_k]$.

To test our hypothesis that utilising context images from target domain can improve domain adaptation, we propose \emph{contextual embeddings}. We condition the inference on $I_k$ by embedding $C_v(I_k)$ at the output of the encoder $E[I_k]$ as follows:  $C_e[I_k]  = E[I_k] \circledast C_v[I_k]$, where $\circledast$ is the embedding operator. The decoder works on the embedding output and produces output prediction $\hat{S}^\prime_{K} = D \circ C_e[I_k]$. Finally, we optimize for loss between $\hat{S}^\prime_{k} $ and ground truth shape $S_k$ using $J(\theta)$. We denote the learnt model as $\mathcal{\theta}^{cn}$ as \emph{ContextNets}
which maps an input image and its context vector to output predictions - $\mathcal{\theta}^{cn} : \{I_k, C_v\} : \rightarrow S_k$.

\section{ContextNets for X-ray Lung Segmentation}
\label{Exp}
In this section, we show an application of $ContextNets$ on the challenging X-ray segmentation problem, by studying 3 independent cohorts varying in disease type, intensity patterns \& contrast:

\begin{itemize}
\item{\textbf{Montgomery dataset}: \cite{jaeger2014two} The Montgomery TB dataset is an open source dataset consisting of 138 posterior-anterior X-rays, of which 80 X-rays are normal and 58 X-rays are abnormal with manifestations of tuberculosis.}
	 \label{Mont}

\item{\textbf{JSRT dataset}: \cite{shiraishi2000development} : The JSRT database consists of 274 chest X-rays with lung masks collected from typical clinical practice, but with differing image distributions.}
% \vspace{- 0.07in}

\item{\textbf{Pneumoconiosis dataset} : Pneumoconiosis is an occupational disease characterized by settlements of dust and other metal particles in the lungs. This internal dataset consists of 330 images acquired from a hospital site with high incidence of pneumoconioisis.}
\end{itemize}
We describe our design choices for the image and context features for X-ray Lung segmentation problem. We used two sets of context features - to model texture and
shape respectively.

{\textbf{Image Features}}: We explored  image similarity features from computer vision like wavelet features, histogram of gradients (HoG) among others. We also explored features extracted from popular pretrained models  like VGGNet, ResNet, etc. We evaluated the clustering similarity qualitatively and  observed that the context set $C(I_k)$  derived using  discrete Haar wavelet features was satisfactory.

{\textbf{Texture Features}: We chose to use features extracted from fully connected layer (fc1) of the popular VGGNet architecture as our texture descriptors, as they have been shown effective on multiple limited data problems.  The intuition is that, texture features from context set should help in handling cases with unexpected texture/contrast variations.

\textbf{Shape Features}: Inspired by \cite{ravishankar2017learning}, we built a light-weight shape auto-encoder (SAE), which takes shape masks as input and projects it to the manifold of training shapes.  We extract features at the ouput of encoder of SAE, with feature length  = $256$. Constraining the inference of a test image on the shape properties of the context set,  can add robustness and regularization to the predictions.  Additionally, on  cases like pathology or pneumoconiosis or TB, where lungs might be partially affected, having shape descriptors from masks on images from context set help improve the performance.

We proceed to build \emph{ContextNets} from source domain data using procedure described in Sec. \ref{formulation}.

\subsection{\emph{ContextNets} in action}
We explore both supervised and unsupervised domain adaptation variants as the availability of ground truth masks from target domain may not always be available.
Equations \ref{CN1} and \ref{CN2} describe the training procedure for learning \emph{$ContextNet_{1}$} and \emph{$ContextNet_{2}$}. $q, t $ and $g$ are wavelet features for querying, VGGNet features for texture and SAE features for shape, respectively. \emph{$ContextNet_{1}$} utilises the input image ($I_k$) along with texture features of the context set ($T_v(C(I_k))$) from source domain memory  and learns $\theta_1^{cn}$ to determine segmentation masks $S_k$. \emph{$ContextNet_{2}$} additionally utilises shape features ($G_v(C(I_k))$)  to learn $\theta_2^{cn}$. Memory module  $\mathcal{M}^s$ appropriately contains only the texture features for \emph{$ContextNet_{1}$} and both texture and shape features for \emph{$ContextNet_{2}$} -which corresponds to unsupervised \& supervised settings respectively.
\vspace{0.1cm}
\begin{gather}
\theta_1^{cn}\!: \{I_k, T_v(C(I_k)\} \rightarrow \!S_k, \,\;\mathcal{M}^s=\{q, t\}
\label{CN1}
\\
\theta_2^{cn}\!: \{I_k, T_v(C(I_k), G_v(C(I_k)\} \,\!\!\rightarrow \!S_k, \; \mathcal{M}^s=\{q, t, g\}
\label{CN2}
\end{gather}
Finally, for every test image ${I^d_k}$ from a target domain $d$, we fetch context set and context vectors from target domain memory  $\mathcal{M}^d$ and infer the segmentation masks using $\theta_1^{cn}$ or $\theta_2^{cn}$ as shown in Fig. \ref{fig:mann}.
\begin{figure}[h]
    % \centering

    \subfloat[Schematic of ContextNets in action \label{fig:mann}]{
		\centering
    \includegraphics[width=.45\textwidth]{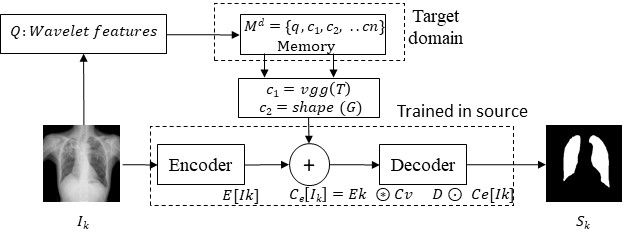}
    }
		\hfill
    \subfloat[Quantitative Results using \emph{ContextNet} \label{table:objective}]{
		\centering
		\includegraphics[width=.45\textwidth]{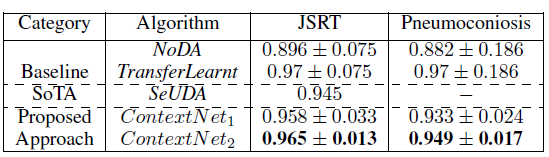}
    }

\end{figure}

\section{Experiments and Results}

We use the Montgomery dataset as the source domain and evaluate the adaptation performance on the other 2 domains. Our FCN architecture contains  $8$ convolutional layers shared equally between the encoder and decoder. We experimented with different combinations like concatenation, average and sum for our context embedding function. The SAE was another fully-convolutional neural network with very few parameters $\sim$ {10k}. We also varied the size of the context from  $3$ to $5$ and obtained similar results. In rest of the paper, we will  share results for context size of $5$ and average of context vectors as our embedding. We optimized for BCE loss and all the models were trained for $100$ epochs with batch\_size = 5.
Further, we perform standard  preprocessing steps like histogram normalization to ensure reasonable baseline comparison, without which the baseline results were very poor.

 We use Dice coefficient between predicted and ground truth masks as the performance evaluation metric. We benchmark  \emph{$ContextNet_{1,2}$} with following approaches:
\begin{itemize}
\item{\textbf{NoDA}: We build U-Net  on the source domain data, and test on target domains without
any adaptation. This scenario establishes the lower baseline for comparisons.}
\item{\textbf{TransferLearnt}: If the entire set of  target domain images and masks are available, the obvious strategy is to retrain/finetune the source domain trained model. This is the gold-standard performance. However, this impractical scenario is included, as it establishes the upper limit performance.}
\item{\textbf{SeUDA}: We compare our algorithm with the state-of-the-art results for unsupervised domain
adaptation, a task-aware generative adversarial network\cite{chen2018semantic}. We however note that
these methods require access to source domain dataset during deployment setting. We report their
results on JSRT from \cite{chen2018semantic} after averaging the right and left lung performance. }

\end{itemize}

From Table \ref{table:objective}, it is apparent that \emph{NoDA} suffers immensely
when subjected to domain changes. On JSRT dataset, both \emph{$ContextNet_{1,2}$} show remarkable improvements of 6$\%$ and 7$\%$, respectively over \emph{NoDA}. Interestingly, our algorithm \emph{$ContextNet_{2}$} even outperforms state-of-the art method - \emph{SeUDA} which requires access to the source dataset, along with additional computational burden of building GAN models per target domain. Table \ref{table:objective} also demonstrates the efficacy of proposed approach on the challenging pneumoconiosis dataset which contain lungs that are partially/fully affected by the disease.
With no retraining or requirement of source/target domain data, we demonstrate remarkable performance and almost match the impractical upper baseline - \emph{TransferLearnt}.

\begin{figure}[!htbp]
\centering
\resizebox{0.8\textwidth}{!}{%
\begin{tabular}{cccccc}
\raisebox{8\height}{JSRT example} &
\includegraphics[width=0.23\columnwidth]{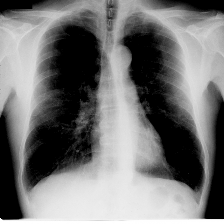} &
\includegraphics[width=0.23\columnwidth]{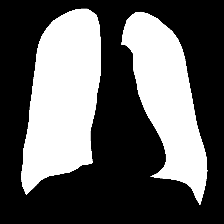} &
\includegraphics[width=0.23\columnwidth]{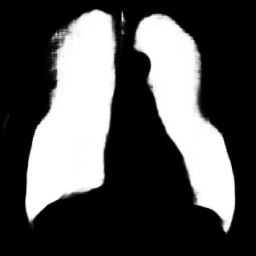} &
\includegraphics[width=0.23\columnwidth]{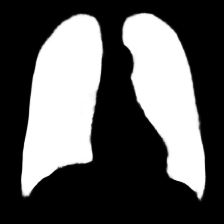} &
\includegraphics[,width=0.23\columnwidth]{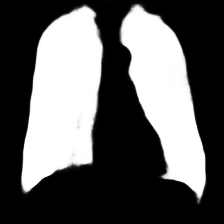} \\

\raisebox{8\height}{JSRT female scan} &
\includegraphics[width=0.23\columnwidth]{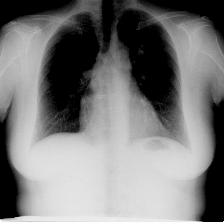} &
\includegraphics[width=0.23\columnwidth]{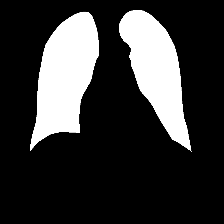} &
\includegraphics[width=0.23\columnwidth]{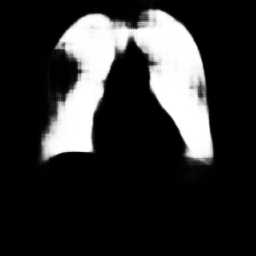} &
\includegraphics[width=0.23\columnwidth]{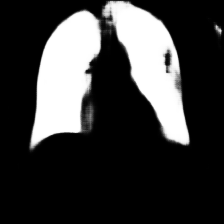} &
\includegraphics[width=0.23\columnwidth]{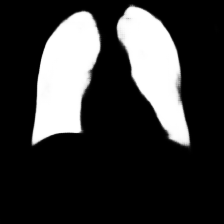} \\

\raisebox{8\height}{Pneumoconiosis} &
\includegraphics[,width=0.23\columnwidth]{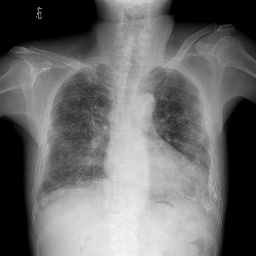} &
\includegraphics[,width=0.23\columnwidth]{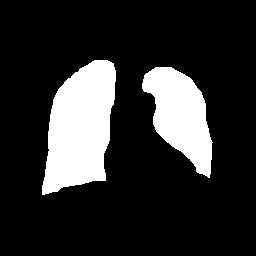} &
\includegraphics[width=0.23\columnwidth]{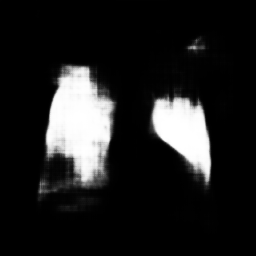} &
\includegraphics[width=0.23\columnwidth]{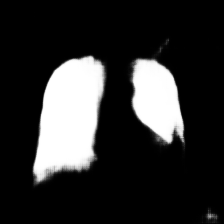} &
\includegraphics[width=0.23\columnwidth]{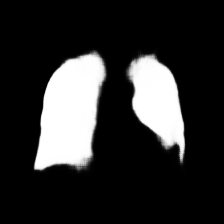} \\

& (a) & (b) & (c) & (d) & (e)

\end{tabular}
}
\vspace{-5pt}
\caption{ Visualization of \emph{ContextNet} results and comparisons on three exemplar target images. Column (b) displays ground-truth masks. Column(c) depicts UNet results. Columns (d) and (e) correspond to \emph{$ContextNet_1$} and \emph{$ContextNet_2$} respectively}

    \label{fig:subjectiveComparison}
\end{figure}

Fig \ref{fig:subjectiveComparison} presents visualization of results. The columns correspond to 3 exemplar images and rows display results from different methods. These images are chosen in increasing order of domain adapation complexity. The first image suffers from texture variations and second image depicts structural variations due to female breasts. The final column shows an extreme case of domain adaptation challenge due to manifestation of stage-3 pneumoconiosis. In all the cases, UNet's performance degrades significantly. Both $ContextNet_{1}$ and $ContextNet_2$ perform similarly on the first case, as the challenge is only due to texture varations. However, on cases 2 and 3, effect of modeling shape plus texture is apparent, with $ContextNet_2$ producing smoother and accurate shape masks compared to $ContextNet_1$. Finally, by limiting the target memory size to just $50$, we tested the robustness of the methods, and obtained similar performance of $\sim$ $0.96$ for $ContextNet_2$ on JSRT dataset.

\section{Conclusion}

In this paper, we propose a practical method for domain adaptation for image segmentation.  We emulate  expert clinicians' process of drawing experience from similar cases through the use of an external memory. Our approach  achieves continuous domain adaptation
with extremely few annotated samples from the target domain. We also alleviate privacy issues by eliminating the need to store source domain data. Further, our method is  a  lifelong learning paradigm  suited to dynamic changes in data distributions. Future research would be focussed on automatic design of various components of the system and a memory management module.

% \bibliographystyle{plain}
% \bibliography{refs}
% \input{supplementary_section.tex }

\end{document}